\documentclass[letterpaper, 10 pt, conference]{ieeeconf-modified}
\IEEEoverridecommandlockouts                              %

\usepackage{graphicx}
\usepackage{amsmath} 
\usepackage{amssymb}

\usepackage{mathtools}

\usepackage[numbers]{natbib}
\usepackage{booktabs}
\usepackage{hyperref}
\usepackage[capitalize]{cleveref}

\usepackage{siunitx}
\usepackage{fancyhdr}

\crefname{equation}{}{}  %

\newcommand{\gitlink}{https://dlr-alr.github.io/2023-humanoids-contact/}

\newcommand{\RescaleSymbol}[2][.75]{\mathop{\vcenter{\hbox{\scalebox{#1}{$#2$}}}}}

\DeclareFontEncoding{LS1}{}{}
\DeclareFontSubstitution{LS1}{stix}{m}{n}
\DeclareSymbolFont{symbols4}{LS1}{stixbb}{m}{it}
\DeclareMathSymbol{\helphexagon}{\mathord}{symbols4}{"DD}

\newcommand{\bcircle}{\RescaleSymbol[1.3]{\bullet}}
\newcommand{\btriangle}{\RescaleSymbol[1.0]{\blacktriangle}}
\newcommand{\bhexagon}{\RescaleSymbol[0.7]{\helphexagon}}

\newcommand{\cov}{\operatorname{cov}}
\newcommand{\diag}{\operatorname{diag}}
\newcommand{\kernel}{\operatorname{kernel}}

\newcommand{\norm}[1]{\left\lVert#1\right\rVert}
\newcommand{\at}[2][]{#1|_{#2}}

\newcommand{\ConfigSpace}{\mathcal{Q}}

\newcommand{\Joints}{q}

\newcommand{\Rot}{\operatorname{Rot}}
\newcommand{\Trans}{\operatorname{Trans}}

\newcommand{\Trobot}{F}
\newcommand{\T}[2]{{}^{#1}T_{#2}}
\newcommand{\Tcamerarobot}{\T{\mathrm{c}}{0}}
\newcommand{\SOThree}{SO(3)}
\newcommand{\RThree}{\mathbb{R}^3}
\newcommand{\ROne}{\mathbb{R}}

\newcommand{\EE}{E}
\newcommand{\EEk}{\EE_k}
\newcommand{\EEl}{\EE_l}

\newcommand{\funForward}{f}

\newcommand{\funMeas}{h}
\newcommand{\funMeasTask}{\funMeas_\mathrm{t}}
\newcommand{\funMeasVicon}{\funMeas_\mathrm{v}}
\newcommand{\funMeasContact}{\funMeas_\mathrm{c}}

\newcommand{\error}{\epsilon}
\newcommand{\errorContact}{\error_{\mathrm{c}}}
\newcommand{\errorVicon}{\error_{\mathrm{v}}}

\newcommand{\CALpar}{\Theta}

\newcommand{\stndDevPrior}{\sigma_{\mathrm{p}}}
\newcommand{\prior}{\Lambda_{\mathrm{p}}}
\newcommand{\CALparNominal}{\CALpar_{\mathrm{p}}}

\newcommand{\stndDevMeas}{\sigma_{\mathrm{m}}}

\newcommand{\prediction}{y}
\newcommand{\JacMeas}{J}
\newcommand{\JacMeasAll}{\mathbf{J}}
\newcommand{\JacMeasTask}{J_\mathrm{t}}
\newcommand{\JacMeasContact}{J_\mathrm{c}}
\newcommand{\JacMeasAllContact}{\mathbf{J}_\mathrm{c}}
\newcommand{\JacMeasAllTask}{\mathbf{J}_\mathrm{t}}

\newcommand{\optimalityD}{O_\mathrm{D}}

\newcommand{\DHpar}{\rho}
\newcommand{\DHd}{d}
\newcommand{\DHtheta}{\theta}
\newcommand{\DHr}{r}
\newcommand{\DHalpha}{\alpha}

\newcommand{\NDoF}{N_{\mathrm{DoF}}}
\newcommand{\NEE}{N_{\mathrm{\EE}}}  
\newcommand{\Nframes}{N_{\mathrm{F}}}
\newcommand{\NMeas}{N_{\mathrm{D}}}
\newcommand{\NMeasTest}{N_{\mathrm{\Bar{D}}}} 
\newcommand{\NTheta}{N_{\Theta}}

\title{\LARGE \bf
Self-Contained and Automatic Calibration of a Multi-Fingered Hand\\Using Only Pairwise Contact Measurements
}
\author{Johannes Tenhumberg$^{* 1, 2, 3}$ \;\; Leon Sievers$^{* 1, 2, 3}$ \;\; Berthold Bäuml$^{1,2}$%
\thanks{$^{*}$ First two authors contributed equally.}
\thanks{$^{1}$DLR Institute of Robotics \& Mechatronics, Germany}
\thanks{$^{2}$Deggendorf Institute of Technology, Germany} 
\thanks{$^{3}$Technical University of Munich, Germany}
\thanks{Contact: \footnotesize johannes.tenhumberg@dlr.de, leon.sievers@dlr.de}
}

\fancypagestyle{FirstPage}{
\lfoot{\fontsize{7}{7}\selectfont \begin{center}\copyright2023 IEEE. Personal
use of this material is permitted. \\Permission from IEEE must be
obtained for all other uses, in any current or future media,
including reprinting/republishing this material for advertising or
promotional purposes, creating new collective works, for resale or
redistribution to servers or lists, or reuse of any copyrighted
component of this work in other works.\end{center}}
}

\begin{document}

\maketitle
\thispagestyle{empty}
\pagestyle{empty}

\begin{abstract}
A self-contained calibration procedure that can be performed automatically without additional external sensors or tools is a significant advantage, especially for complex robotic systems.
Here, we show that the kinematics of a multi-fingered robotic hand can be precisely calibrated only by moving the tips of the fingers pairwise into contact. 
The only prerequisite for this is sensitive contact detection, e.g., by torque-sensing in the joints (as in our DLR-Hand II) or tactile skin. 
The measurement function for a given joint configuration is the distance between the modeled fingertip geometries, but the actual measurement is always zero. 
In an in-depth analysis, we prove that this contact-based calibration determines all quantities needed for manipulating objects with the hand, i.e., the difference vectors of the fingertips, and that it is as sensitive as a calibration using an external visual tracking system and markers.
We describe the complete calibration scheme, including the selection of optimal sample joint configurations and search motions for the contacts despite the initial kinematic uncertainties.
In a real-world calibration experiment for the torque-controlled four-fingered DLR-Hand II, the maximal error of \SI{17.7}{mm} can be reduced to only \SI{3.7}{mm}.\\
Web: \href{\gitlink}{\gitlink}
\end{abstract}

\section{Introduction}\label{sec:Introduction}
\thispagestyle{FirstPage}
Autonomous robots that robustly perform dextrous manipulation tasks in the real world generally require precise models of the system's kinematics. 
Regarding multi-fingered robotic hands, one crucial class is planning algorithms for finding an optimal grasp for a given 3D model of an object~\cite{Winkelbauer2022Grasping, Winkelbauer2022}.
This task depends on the precise placement of the fingers on the object's surface. 
Another class is methods for dextrous in-hand manipulation. 
Here, only recent modern deep reinforcement learning algorithms trained in simulation have enabled dexterity close to human performance.
Primarily when performed in a purely tactile setting~\cite{Sievers2022LearningHand, Pitz2023DextrousArchitecture, Rostel2023}, i.e., without cameras, where only joint angles (and tactile measurements, e.g., via torque-sensing) are used, robust zero-shot sim2real transfer requires a precise kinematics model with a maximal error of a few millimeters.

It is desirable to have a quick calibration procedure that runs entirely automatically and is self-contained, i.e., it does not need any additional external sensors or tools to obtain a precise kinematic model.

\subsection{Related Work}

\begin{figure}[t]
    \centering
	\includegraphics[width=1\linewidth]{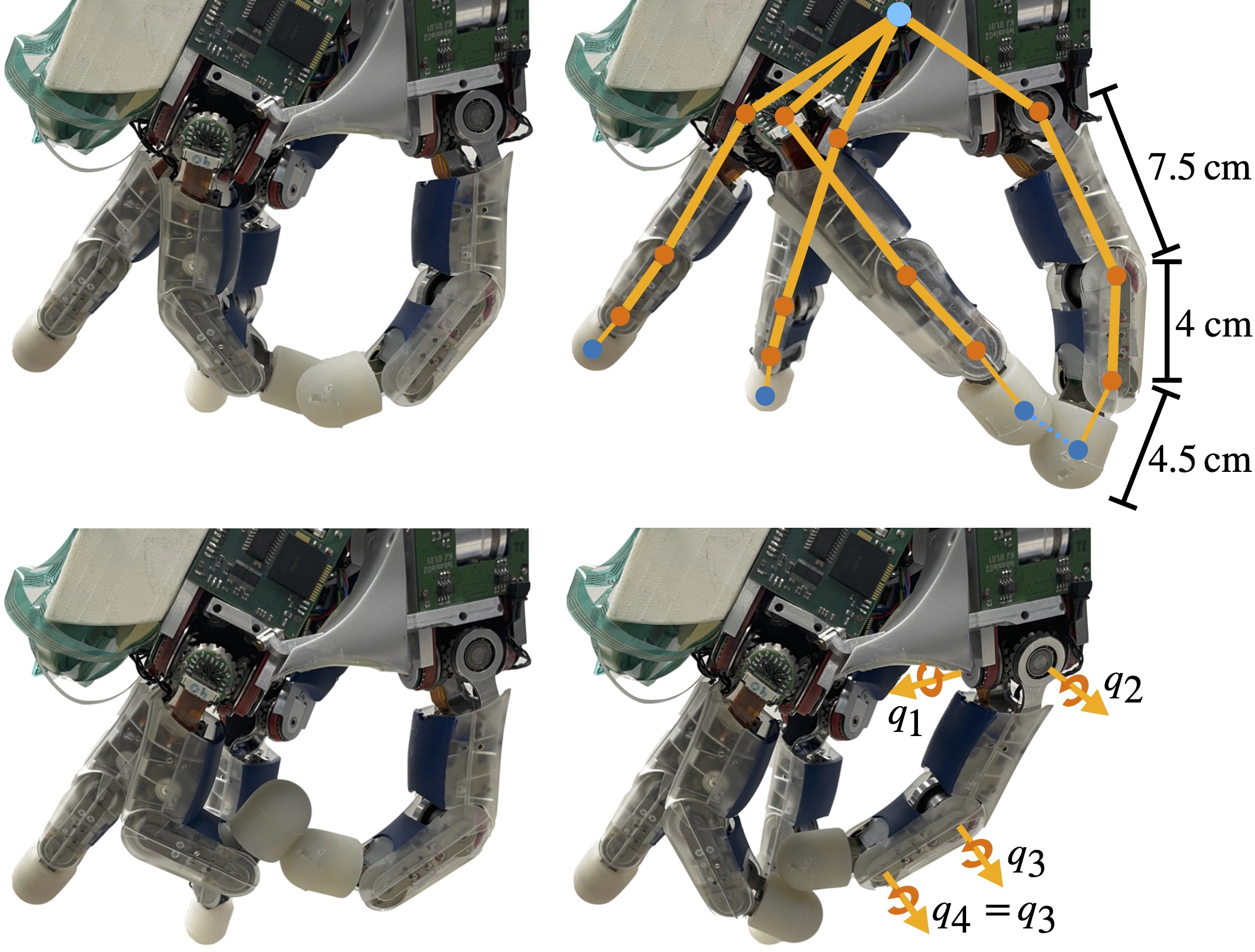}
	\caption{DLR-Hand II~\cite{Butterfass2001DLRHand} with thumb and index finger in contact for different poses.
    The kinematic tree of the whole hand is indicated in orange, and the three active joints plus the fourth passive joint are drawn together with the dimensions of the fingers.
    The other two fingers are far extended to allow for a large shared workspace of the current pair. 
    For the whole calibration, all six finger pairs are measured and calibrated jointly.}
    \label{fig:hand_poses_annotated}
\end{figure}

A classical and often-used approach for calibrating kinematic chains and trees is visual tracking of the end-effector(s).
For a robotic hand, this visual approach with an external tracking system was demonstrated by~\citet{Lee2013HandVision}.
However, adding visual markers to the many end-effectors is pretty cumbersome, especially as more than one marker is usually required per fingertip due to the mutual occlusions of the many fingers in the small workspace of a hand. 
Using an electromagnetic tracking system to measure the position and orientation of the fingertips~\cite{Tan2018HandElectromagnetic} is less prone to occlusions. 
However, the requirement for additional hardware hinders the ease of use of such an approach.

Another calibration procedure that avoids the need for a tracking system uses geometric constraints on the kinematic chain.
An early work constrained the motion of an end-effector on a plane to calibrate a robotic arm \cite{Ikits1997PlaneConstraint}.
Since then, there have been examples of using mechanical fixtures~\cite{Meggiolaro2000EndpointContact}, relative calibration techniques~\cite{Sun2009RelativeTechniques}, and precise reference plates~\cite{Joubair2015KinematicConstraints} to calibrate a robotic arm without a vision system.
\citet{Bennett1990ClosedLoop} applied these ideas to a robotic hand and calibrated the multi-finger Utah-MIT hand by rigidly connecting two fingertips with a plate, resulting in a closed-loop kinematic chain. 
A downside of all those approaches is that the mounting procedure can damage the fingers and takes time, especially if it needs to be repeated for each finger pair.

A particular case of the relative geometric calibration techniques does not require mechanical fixtures to enforce the constraints but relies on self-contact.
The humanoid robot iCub was intensely used for such studies. 
First in simulation~\cite{Stepanova2019SimulationICUB} and then on the actual hardware together with other sensing modes~\cite{Stepanova2022MultipleSensors}.
\citet{Roncone2014SkinTouch} calibrated the full DH parameters of the iCub robot using self-touch with a tactile skin with 4200 sensing points. 
Besides identifying the parameters of a given kinematic chain, self-contact can also be used to incrementally update the kinematic scheme~\cite{Zenha2018SchemaAdaption} or identify the layout of an artificial tactile skin~\cite{Rustler2021SpatialCalibration}.

\subsection{Contributions}
In prior work, an external camera system was first used to calibrate an elastic kinematic model of the humanoid Agile Justin~\cite{Tenhumberg2021Elastic}. 
Then, this approach became more accessible by using the robot's internal RGB camera to calibrate the complete kinematic tree~\cite{Tenhumberg2022RGB}. 
This paper extends the calibration to the DLR-Hand II~\cite{Butterfass2001DLRHand, HaiyingHu2004CalibratingHand} (see \cref{fig:hand_poses_annotated}) of the robot and introduces a contact-based approach that does not rely on anything other than the kinematic tree structure itself.
The main contributions of this work are:
\begin{itemize}
\item We show, to our knowledge for the first time, that the kinematics of a multi-fingered hand can be calibrated using only contact measurements of (all) finger pairs, i.e., without any external tools like cameras or markers.
\item An in-depth theoretical analysis of the calibration problem with redundant parameters, including proof that the quantities relevant for manipulation tasks, i.e., the difference vectors between the fingertips, are entirely determined by our contact measurement scheme.
\item A comparison of calibration by contact measurements with calibration using a visual tracking system (theoretical analysis and simulation experiments).
\item A complete scheme for executing a contact-based calibration for a general multi-fingered hand, including optimal selection of finger configurations and planning of search motions to deal with the initial uncertainties in the kinematics.
\item Real-world experiments with the contact calibration of the DLR-Hand II resulting in a reduction of the maximal error from \SI{17.7}{mm} for the nominal kinematic to only \SI{3.7}{mm} for the full DH parameters calibration.
\end{itemize}

\section{Robot Model}\label{sec:RobotModel}
The forward kinematics of a robot maps from the configuration space of generalized joint angles to the robots' physical pose in the cartesian workspace.
Both for robotic arms and hands, an accurate model of the robot's kinematics is essential for precise and collision-free motions.
A well-fitted model is also required to perform challenging grasping and manipulation tasks.
 
DH parameters are widely used to describe this kinematic model of the robot.
In this formulation, four values $\rho_i= [\DHd_i, \DHr_i, \DHalpha_i,\DHtheta_i]$ describe the connection between two consecutive frames of the robot:
\begin{align}
\T{i-1}{i} =
\Rot_{\mathrm{x}}(\DHalpha_{i}) \cdot
\Trans_{\mathrm{x}}(\DHr_{i}) \cdot
\Rot_{\mathrm{z}}(\DHtheta_i) \cdot
\Trans_{\mathrm{z}}(\DHd_i)
\label{eq:DH2Frame} 
\end{align}
The joints $\Joints_i$ are treated as offsets to $\DHtheta_i$ or $\DHd_i$ in \cref{eq:DH2Frame} depending on the type of joint.
This minimal representation with two translational and two rotational parameters is enough to describe an arbitrary robot.

The frame of the end-effector ($i=\EE$) relative to the robot's base ($i=0$) is calculated by applying the transformations in series:
\begin{align}
\T{0}{\EE}  = \T{0}{1} \cdot \T{1}{2} \cdot \dotsc \cdot \T{\EE-1}{\EE}
\label{eq:TTT}
\end{align}
For robots with a kinematic tree structure with multiple end-effectors $\EEk, k=1\ldots\NEE$, equation \cref{eq:TTT} holds for each branch.
The forward kinematics maps from joint configurations $\Joints \in \ConfigSpace$ to all the $\Nframes$ frames of the robot 
\begin{align}
\funForward(\Joints, \DHpar) = \Trobot = [\T{0}{1}, \T{0}{2}, \dotsc, \T{0}{\Nframes}].
\end{align}
Each frame $\Trobot_i$ describes a full 6D pose $(\Trobot_{i, x}, \Trobot_{i, r}) \in \RThree \times \SOThree$.

\begin{figure}[t]
    \centering
	\includegraphics[width=1\linewidth]{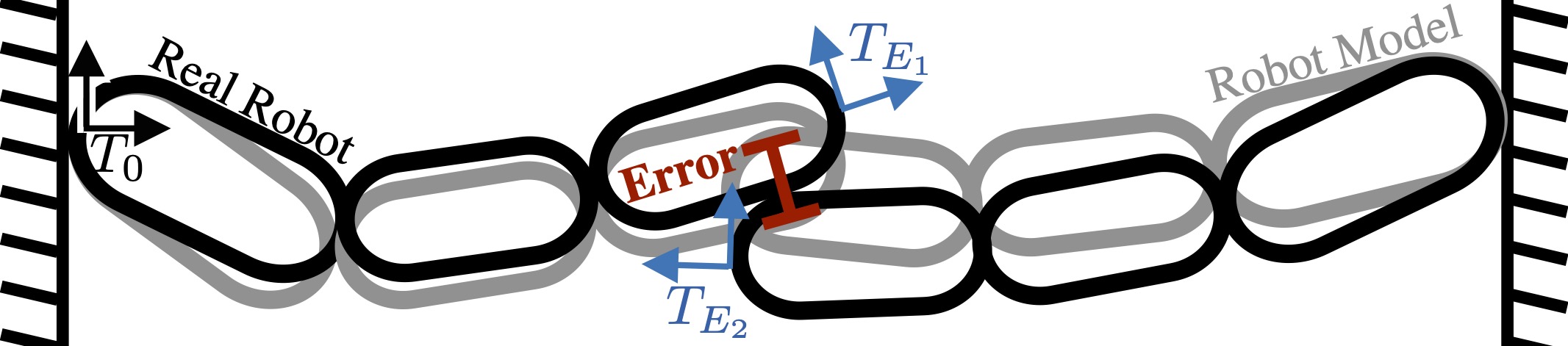}
	\caption{The scheme shows the actual contact measurement on the hardware (black) and the same joint configuration applied to the robot model (gray). 
    The fingers are in penetration for the current set of calibration parameters. 
    This error (red) should be minimized over the calibration process.}
    \label{fig:contact_measurement_model_vs_true}
\end{figure}

\section{Calibration for In-hand Manipulation}\label{sec:Hand-Calibration}

\subsection{Identification}
The goal of the calibration is to identify this kinematic model. 
In our case, the central parameters of the forward model are the DH parameters $\DHpar$.
Nevertheless, in general, there can be additional parameters that have to be estimated jointly with the kinematic parameters. 
Examples are camera intrinsics or additional frames to close the measurement loop.
We denote the combination of all calibration parameters $\CALpar$.
These parameters of the measurement function $\funMeas$ can be identified using a dataset $D = \{(\Joints^{(n)}, y^{(n)})\}_{n=1\ldots N}$ of corresponding pairs between the robot's configuration $\Joints^{(n)}$ and the measurement $y^{(n)}$.
We formulate the identification as a single combined least-squares problem based on all measurements and all body parts to minimize the error $\error = \prediction - \funMeas(\Joints, \CALpar)$ between measurements and model predictions.
To find the optimal parameters $\CALpar^*$, we use the maximum a posteriori (MAP) approach:
\begin{align*}
\min_{\CALpar} \left[ \sum_n^N \frac{1}{\stndDevMeas^2} \norm{\prediction^{(n)} - \funMeas(\Joints^{(n)}, \CALpar)}^2 + \Delta \CALpar^T \prior^{-1} \Delta \CALpar \right] \\
\mathrm{with} \; \Delta\CALpar = \CALpar - \CALparNominal
\end{align*}
This approach employs a Gaussian distribution beforehand, characterized by a mean $\CALparNominal$ and a diagonal covariance matrix $\prior = \diag{\stndDevPrior^2}$. 
The uncertainty of the calibration parameters is modeled via $\stndDevPrior$. 
Furthermore, we assume the usual Gaussian distribution with a mean of zero and a standard deviation of $\stndDevMeas$ for the measurement noise.
By incorporating the prior, this method is regularized, ensuring a minimum exists, even if there are redundancies in the measurement or kinematic model.

\subsection{Task Measurement Function}\label{sec:TaskMeasFun}

\begin{figure}[t]
    \centering
	\includegraphics[width=1\linewidth]
    {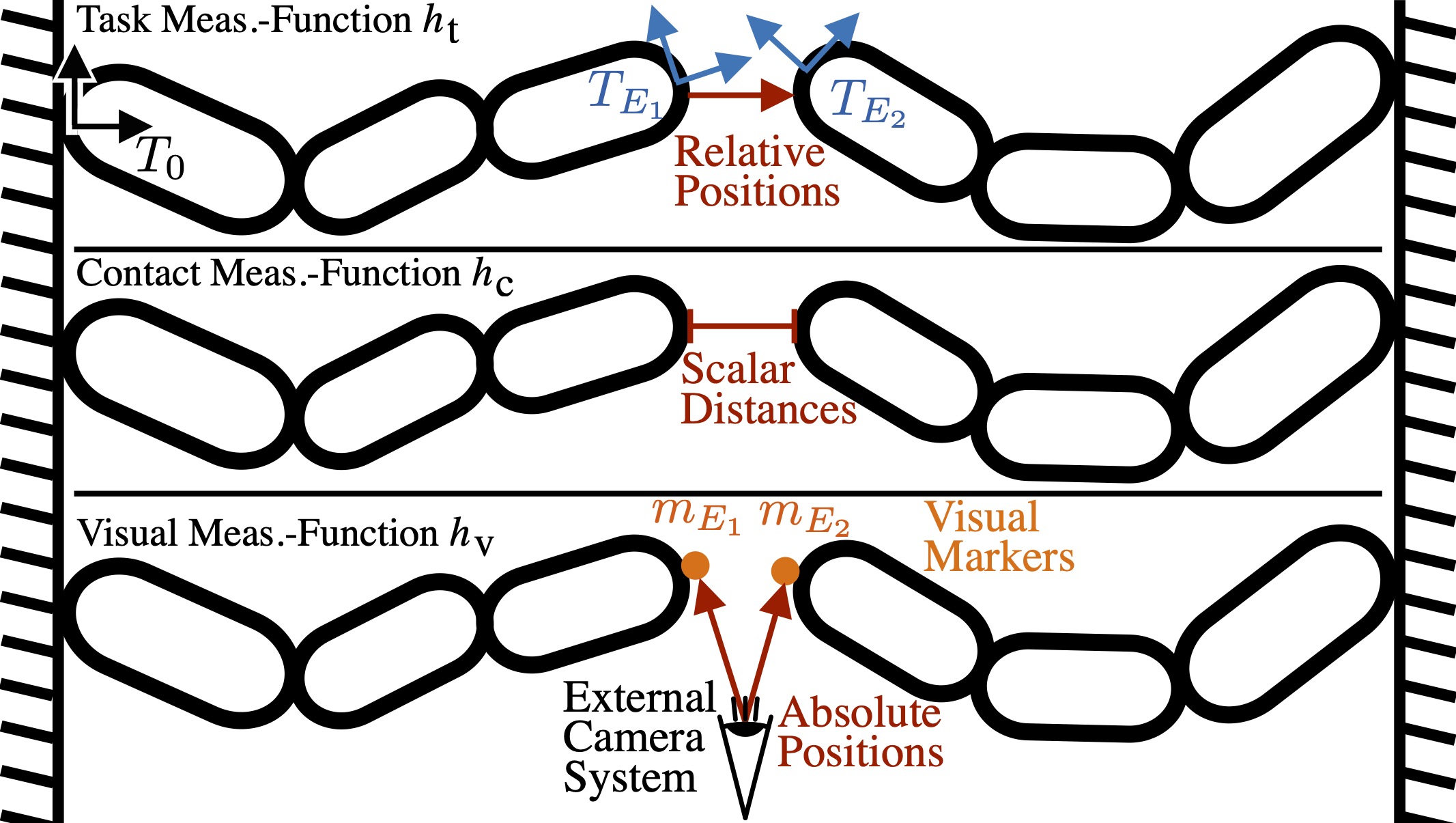}
	\caption{The scheme shows the three measurement functions discussed in \cref{sec:Hand-Calibration}.
    \emph{Top (\cref{sec:TaskMeasFun}):} The task measurement function $\funMeasTask$ measures the relative positions of the end-effectors.
    \emph{Middle (\cref{sec:ContactMeasFun}):} The contact measurement function $\funMeasContact$ measures the scalar distance between two end-effectors.
    \emph{Bottom (\cref{sec:VisMeasFun}):} The cartesian measurement function $\funMeasVicon$ uses an external tracking system to measure the absolute position of the end-effectors.
    }
    \label{fig:measurement_functions}
\end{figure}

We aim to calibrate this kinematic model of the hand for dextrous in-hand manipulation. 
To move the fingers in a controlled manner, as demonstrated by \citet{Pitz2023DextrousArchitecture} with the cube, an accurate model for the relative positions of the fingertips to each other is necessary.
To relate closely to this task, we chose our desired measurement function to measure the relative positions directly. 
The calibration process should then minimize the error of this function.

For $\NEE$ fingers, $\NEE-1$ distance vectors define the whole set. 
One can choose one fingertip $E_1$ as the basis and compute the positions relative to it for the remaining fingers
\begin{align}
y^k = \funMeasTask^k(\Joints, \DHpar) = \funForward(\Joints, \DHpar)_{\EEk, x} - \funForward(\Joints, \DHpar)_{\mathrm{E_1}, x}  \label{eq:funMeasTask}
\end{align}
One can then concatenate the measurements for the pairs to get the combined measurements ${ \funMeasTask = [\funMeasTask^2, \ldots, \funMeasTask^{\NEE}], \funMeasTask^k \in \RThree }$.
This results in $3 \cdot (\NEE-1)$ data points per robot pose for the three spatial directions and $\NEE$ end-effectors.

Note that \cref{eq:funMeasTask} is our theoretically desired measurement function, which measures all the information we care about for precise in-hand manipulation. 
For example, we can not detect a translational offset of all end-effectors like a cartesian tracking system could. 
However, such a shift does not change the fingers' relative behavior and is irrelevant to our task.

\subsubsection*{Task Test Set}\label{sec:TaskTestSet}
Besides the task measurement function, the test set is central to evaluating the calibration quality.
We want a high accuracy across the whole cartesian workspace.
Therefore, the test set for evaluating all our experiments should be uniformly distributed in the cartesian workspace. 
We sample random joint configurations for each finger and map them to the end-effector positions via the forward kinematics $\funForward$. 
In the cartesian workspace, we create a fine grid and draw one of the grid cells uniformly, and in the second step, we draw one joint configuration that lies in this grid cell.
This sampling strategy ensures a uniform distribution over the workspace, which does not hold for configurations sampled randomly in the configuration space.

After defining a suitable measurement function and test set for our task, we discuss the actual measurement methods that can be applied to the hardware.
First, an external camera system that tracks the cartesian position of visual markers followed by our pairwise contact measurements.
Note that there are also additional considerations, especially the independence of particular infrastructure and the low complexity of the measurement setup, speaking in favor of contact-based measurements.

\subsection{Cartesian Measurement Function}\label{sec:VisMeasFun}
One option to perform the actual measurements is to mount visual markers on the kinematic tree and use an external tracking system to collect cartesian measurements of these markers. 
Assuming that the markers are fixed at position $m_{\EEk}$ relative to end-effector $\EEk$, the cartesian measurement function
\begin{align}
y^k = \funMeasVicon(\Joints, \CALpar)^k = \Tcamerarobot \cdot \funForward(\Joints, \CALpar)_{\EEk} \cdot m_{\EEk} \label{eq:funMeasVicon}
\end{align}
describes how a marker moves dependent on the joint configuration $\Joints$ and the calibration parameters $\CALpar$.
Suppose one wants to calibrate the forward kinematics jointly for multiple end-effectors. In that case, one can combine the measurements for each of the $\NEE $ markers to a combined measurement function $ \funMeasVicon = [\funMeasVicon^1, \ldots, \funMeasVicon^{\NEE}], \funMeasVicon^k \in \RThree $.
While, in general, the measurements with an external camera can be made without constraining the robot directly, one still needs to account for self-collision and a clear view of the markers. 
These additional constraints reduce the possible space in which measurements can be collected.

\subsection{Contact Measurement Function}\label{sec:ContactMeasFun}
\begin{figure}[t]
    \centering
	\includegraphics[width=0.8\linewidth]
    {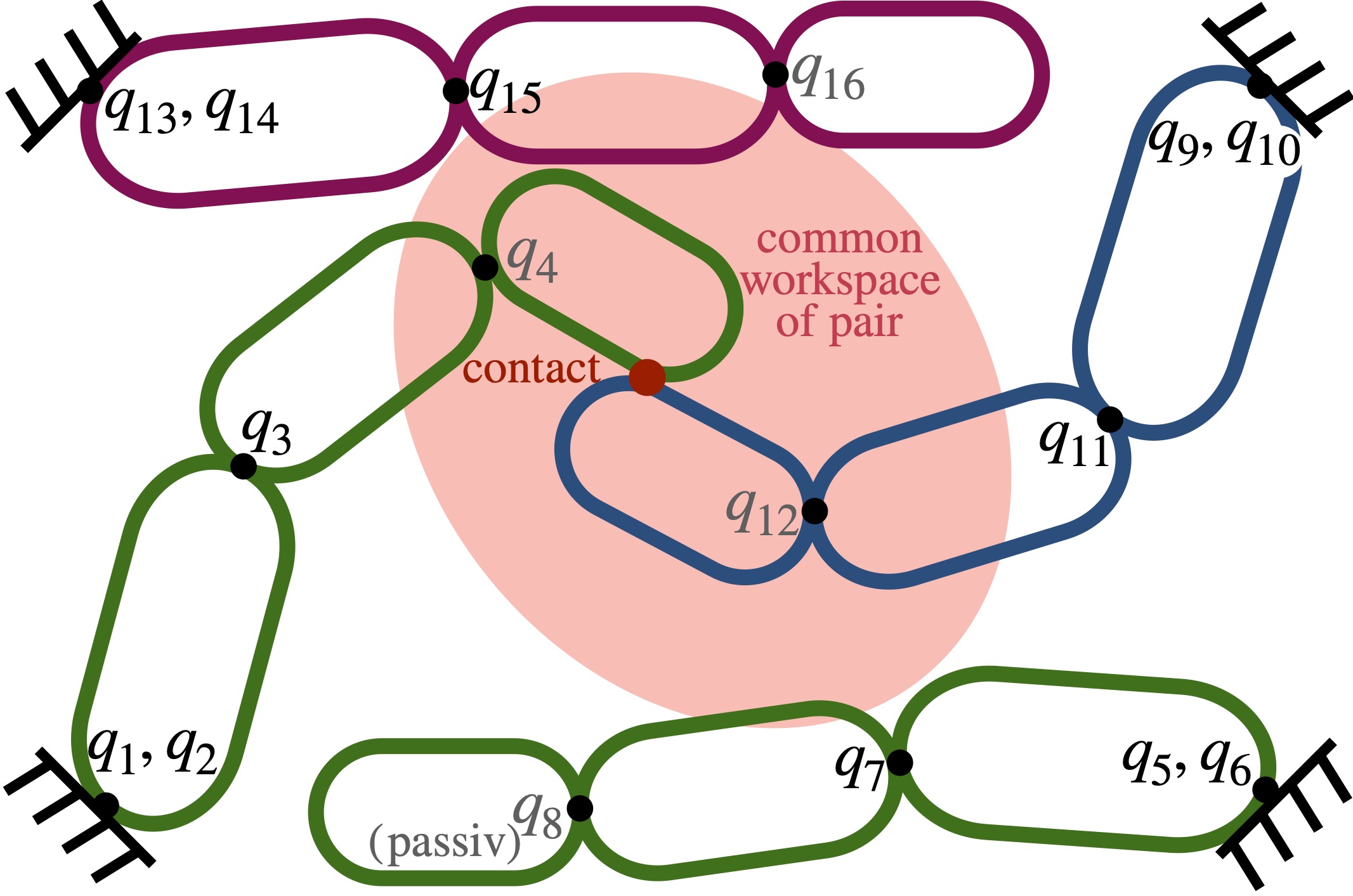}
	\caption{The graphic shows the contact calibration approach for a four-fingered hand.
    For pairwise contact, two fingers always need to move out of the shared workspace of the current finger pair to ensure that self-collisions are avoided and the available configuration space for contact detection is well used.}
    \label{fig:contact_calibration_scheme}
\end{figure}

Another option to perform the actual measurement is to use contact information.
We describe this procedure in more detail in \cref{sec:ContactMeasurementProcedure}. 
In general, the corresponding measurement function measures the distance between parts of the robot.
One needs the exact geometry of the two bodies to compute the distance $d^u$ between a pair of bodies $u=(\EEk, \EEl)$ on the kinematic tree.
The distance is only dependent on the relative position and orientation of the bodies
\begin{align}
d^u = d^u(\T{\EEk}{\EEl}) = d^u(\funForward(\Joints, \DHpar)_{\EEk}^{-1} \cdot \funForward(\Joints, \DHpar)_{\EEl}) = d^u(\Joints, \DHpar). \label{eq:measurement_contact}
\end{align}
This relative frame $\T{\EEk}{\EEl}$ is directly computable from the forward kinematics $\funForward(\Joints, \DHpar)$.

If the bodies at $\EEk$ and $\EEl$ have simple geometric forms, one can directly compute the distance $d_m$. 
In the case of the DLR-Hand II, the fingertips are perfect capsules in the contact area and, therefore, straightforward to compute. 
The more general case is that the geometries are given as arbitrary meshes. 
In this case, one has to use, for example, algorithms like GJK to compute the distance.

The contact measurement function for the pair $u$ can now be written as 
\begin{align}
y^u = \funMeasContact(\Joints, \CALpar)^u = d^u(\funForward(\Joints, \DHpar)).  \label{eq:funMeasContact}
\end{align}
This function can only measure the scalar distance between two body pairs. 
With $\NEE$ end-effectors on the kinematic tree, there are in total $N_{\mathrm{U}} = \binom{\NEE}{2}$ pairs.
The combined measurement function is in this case $\funMeasContact = [\funMeasContact^1, \ldots, \funMeasContact^{N_{\mathrm{U}}}], \funMeasContact^k \in \ROne $

The contact measurement adds hard constraints to the data collection (see \cref{sec:SampleGeneration}). 
As only configurations in contact can be measured, the available configuration space is drastically reduced.
Therefore, the remaining subspace can be quite different from the cartesian task test set described in \cref{sec:TaskTestSet}. 
The following section discusses how those different measurement functions relate and how many parameters they can identify.
Furthermore, we discuss the influence of the different calibration and test sets on the sensitivity.

\begin{figure}[t]
    \centering
	\includegraphics[width=1\linewidth]{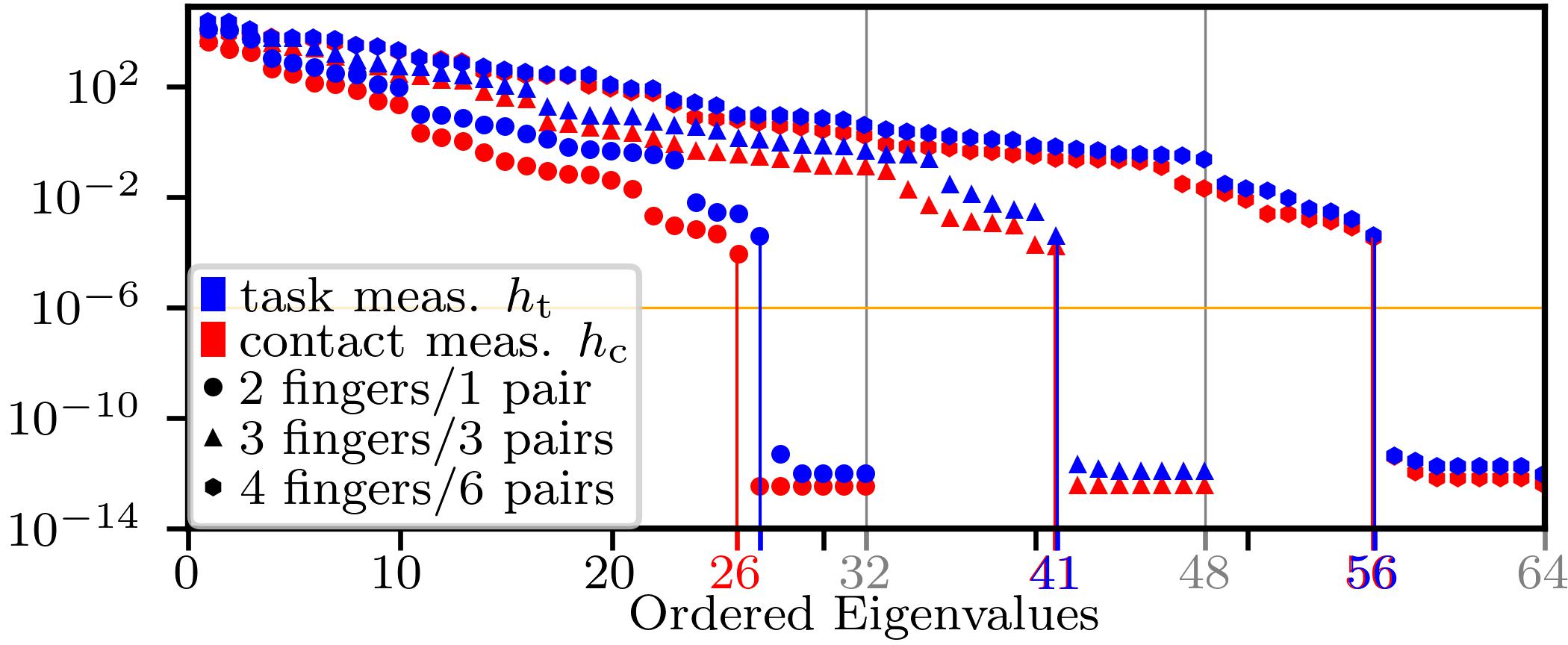}
	\caption{This figure shows the ordered eigenvalues for different measurement setups to analyze the sensitivity. 
    The evaluation was done with the nominal kinematic of the DLR-Hand II; for a more generic hand, see the right plot.  
    The task measurement function $\funMeasTask$ is blue, and our contact measurement function $\funMeasContact$ is red. 
    Furthermore, we show three modes.
    For the one where all the pairs are calibrated simultaneously ($\bhexagon$), the kernels of both measurement functions have the same size. 
    The same is true for the calibration with three fingers ($\btriangle$)
    However, the kernel sizes differ when just a single pair ($\bcircle$) is considered.
    The light gray vertical lines indicate the maximal number of parameters for each mode.
    }
    \label{fig:sensitivity_nominal}
\end{figure}

\begin{figure}[t]
    \centering
	\includegraphics[width=1\linewidth]{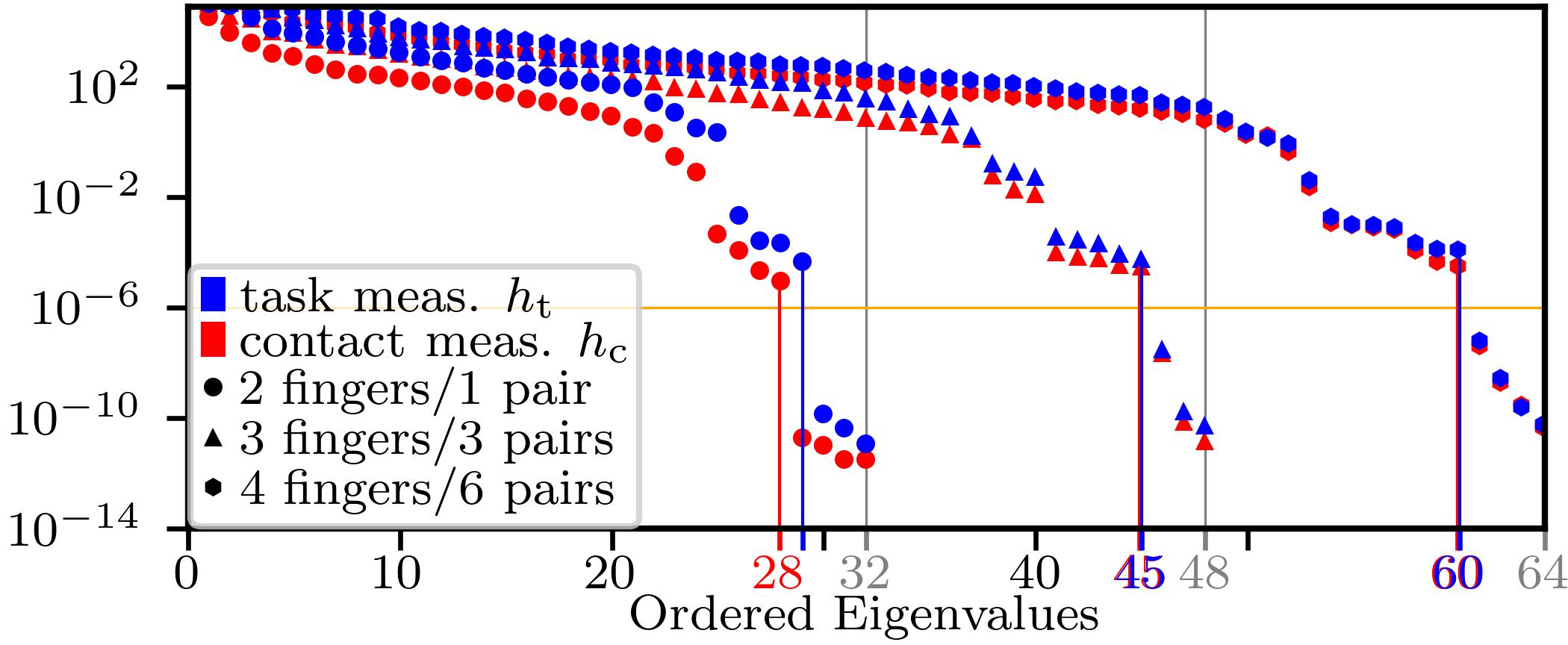}
	\caption{The plot is structured equivalently to \cref{fig:sensitivity_nominal} but for a more generic hand kinematic without parallel axes or mounting frames.
    Here, in theory, all 64 parameters for the entire hand ($\bhexagon$) can be identified. 
    However, when only two fingers are considered ($\bcircle$), the contact measurement function $\funMeasContact$ is still less sensitive than the task measurement function $\funMeasTask$.
    This speaks in favor of a holistic calibration of the whole kinematic tree.}
    \label{fig:sensitivity_generic}
\end{figure}

\section{Problem Analysis}\label{sec:ProblemAnalysis}
\subsection{Sensitivity Analysis}\label{sec:SensitivityAnalysis}

A single contact measurement \cref{eq:funMeasContact} measures less information than a cartesian tracking system \cref{eq:funMeasVicon}. 
The tracking system can measure the absolute position of the markers in the workspace. 
The contact can only measure the relative distance between two parts of the kinematic chain. 
In other words, the contact can measure the joint angles where the distance between the two bodies is zero.

The question is whether this reduced information in the measurement function leads to non-identifiable parameters.
We can answer this question directly by looking at the jacobians of the different measurement functions
\begin{align}
\JacMeas^s = \frac{ \partial \funMeas(\Joints^s, \CALpar) } {\partial \CALpar} \at[\Big]{\CALpar_0}.
\end{align}
Assuming that each measurement is d-dimensional, concatenating those matrices for each measurement leads to the combined jacobian $\JacMeasAll=[\JacMeas^1, \ldots, \JacMeas^{\NMeas}]$ with dimensions $(\NMeas\cdot d) \times \NTheta$.
We can investigate which parameters are identifiable by the measurement function $\funMeas$ by looking at the nullspace of $\JacMeasAll^T \JacMeasAll$. 
The size of the nullspace marks how many of the model parameters $\CALpar$ can not be identified. 
Furthermore, from the eigenvectors corresponding to the eigenvalues close to zero, one can identify the parameters (or sets of parameters) which can not be measured.

When one deals with two distinct measurement functions, as we described with the desired task $\funMeasTask$ and our actual contact measurement function $\funMeasContact$, the necessary condition is that 
\begin{align}
  \kernel(\JacMeasAll_c^T \JacMeasAll_c) \subseteq \kernel(\JacMeasAll_t^T \JacMeasAll_t).
\end{align}
If this condition is satisfied, one can identify all the relevant parameters to the task, defined by $\funMeasTask$.
Note that this is less strict than demanding that both kernels are zero and applies in general to distinct measurement functions for calibration and evaluation.
We allow for unidentifiable parameters if they do not influence our desired measurement function.

\subsection{Optimal Experimental Design}\label{sec:OED}
Following \citet{Carrillo2013TaskOriented}, we use task D-optimality to select appropriate samples for measuring. 
However, we have two key differences in our setup.
First, we have a theoretical desired task measurement function $\funMeasTask$ and an actual measurement function $\funMeasContact$ we can apply on the hardware. 
Ultimately, we want a good fit for the desired measurement function.
Second, the contact measurement reduces the configuration space quite drastically. 
Still, we want a good fit and high accuracy across the whole workspace. 
Therefore we also have two sets here. 
The desired test set is uniformly sampled across the whole cartesian test set and one actual test set, which can be measured via contact detection.
We generalize the optimality framework to account for those mismatches between measurement functions and distributions.
\citet{Carrillo2013TaskOriented} showed that the central equation decouples and the task D-optimality can efficiently be computed with 
\begin{align}
    \optimalityD = \frac{1}{l} \det \big( \cov (\CALpar) \sum_{s=1}^{\NMeasTest} \JacMeasTask^s{}^T \JacMeasTask^s \big). 
    \label{eq:optimalityD}
\end{align}
The mean over $\JacMeas_i^T \JacMeas_i$ is constant for a given test set of size $\NMeasTest$ and a desired task measurement function $\funMeasTask$.
The covariance over the calibration measurements can be estimated using the actual chosen measurement function and the specific calibration set. 
Let $ S = \{ s_i \}_{i=1}^{\NMeas}$ be a subset of a larger calibration set and $\JacMeasAllContact  = [\JacMeasContact^{s_1}, \ldots, \JacMeasContact^{s_{\NMeas}}] $ the combined jacobians corresponding to those measurements.
Then the covariance is given by
\begin{align}
    \cov (\CALpar) = \JacMeasAllContact^{T} \diag(\stndDevMeas^2) \JacMeasAllContact + \diag(\stndDevPrior^2)
    \label{eq:covariance}
\end{align}
and transforms the uncertainty in the measurements $\stndDevMeas$ and the priors $\stndDevPrior$ into an uncertainty in the parameters.

\cref{eq:optimalityD} lets us compute how well different calibration sets are suited to minimize the error of the desired measurement function over a desired test set.    
This criterion can be used to choose a good set of suitable poses for measuring.
Besides reducing the overall size of the necessary calibration set, this selection criterion also counteracts the mismatch in the measurement functions and the calibration and test sets.
We report the results of the sensitivity analysis and the optimal selection criterion in \cref{sec:SimulationStudy}

\section{Contact Measurement Procedure}\label{sec:ContactMeasurementProcedure}
\subsection{Sample Generation}\label{sec:SampleGeneration}

\begin{figure}[t]
    \centering
	\includegraphics[width=1.0\linewidth]{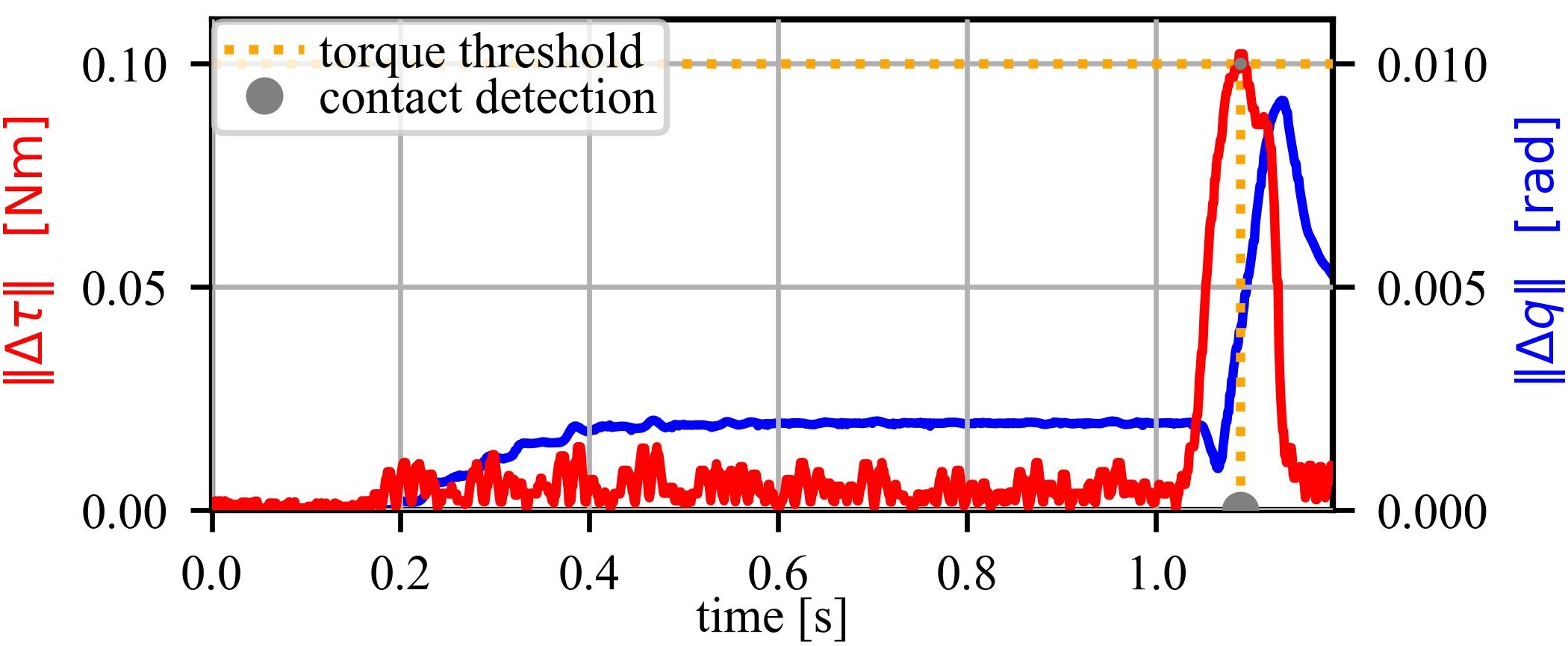}
	\caption{Red: $L^2$ norm of $\Delta\tau$. $\Delta\tau$ denotes the difference between the torque signal measured by the passive finger before the approach started, $\tau_0$, and the currently measured torque.
    Blue: $L^2$ norm of $\Delta q$. $\Delta q$ denotes the difference between the position signal measured by the passive finger before the approach started and the currently measured position.
    The contact is detected when the torque threshold $\tau_t=\SI{0.1}{Nm}$ is exceeded.
    Note how the passive finger's joint angles can change after the active finger starts to drive causing vibrations in the system. The torque sensors only measure noise until the contact occurs.
}
    \label{fig:pos_torque_contact}
\end{figure}

Our approach is to collect pairwise contact measurements for the tree structure of the 
hand. 
A key difference for contact-based calibration is that one does not know the exact joint configuration, which will be measured beforehand.
Generally, the measurement process collects pairs $(\Joints, y)$ to calibrate a function $g(\Joints, \theta) = y$. 
For vision-based measurements, one collects the cartesian position $y_i$ for selected joints $\Joints_i$.
On the other hand, for contact-based measurements, the $y_i$ is known a priori; the contact is, by definition, $y_i = 0$. 
The contact measurement delivers the exact configuration $\Joints_i$, which leads to contact. 

Therefore, contact measurement must also include a search to find the exact configuration in which the contact happens. 
We tackle this problem by generating for each measurement point a trajectory along which contact will probably happen.
We define this path by its endpoints. 
The start configuration is far from contact with the nominal robot model, and the end configuration is deep in penetration. 
Between those endpoints, we then detect contact for the specific finger pair.  

Our approach to generating these search trajectories for the contact measurements consists of multiple steps.
The first step is to find the volume in the workspace which both end-effectors can reach.
We save a large (n=100000) set of configurations that fall in this intersection for both fingers.
The next step is to randomly choose one configuration for finger A and check for which configurations of finger B the tips collide.
The threshold for collision one chooses here strongly depends on the robot model's initial uncertainty. 
Continuing from this pair of configurations, we choose which finger should be static and which should move into contact. 
For the moving finger, we then sample an additional random configuration as the start point of the measurement drive.

We repeat this procedure for all six finger pairs.
For each pair, the rest of the kinematic structure should move as far away as possible from the combined workspace of the current end-effector pair (see \cref{fig:contact_calibration_scheme}).
The goal is to obstruct the measurements as little as possible and allow for diverse joint configurations in this constrained setting.

One additional problem is the small form factor of a robotic hand, especially compared to the errors of the uncalibrated system.
For a robotic arm with a total reach of one meter, an error of a few centimeters does not change the measurement setup. 
However, in our case, the DLR-Hand II with its four fingers, the ratio between error in the forward kinematics and the robot's actual size is much larger. 
We measured uncalibrated errors up to \SI{17.7}{mm}. 
That equals roughly $10\%$ of the workspace and is also about the fingertip size. 
The consequence is that even if, in simulation for the nominal kinematic, the two fingers touch close to the center, it is still possible that the measurement fails on the hardware. 
Therefore one needs to account for this high uncertainty while generating samples and safe trajectories to collect those samples. 
Furthermore, robust contact detection is crucial for reliable measurements on the actual hardware.

\subsection{Contact Detection}

It is essential to detect the precise joint angles in which the contact force between the fingertips is as small as possible.
The DLR-Hand II is equipped with an output side torque sensor for each of the twelve active degrees of freedom.
Before the passive finger gets approached, the offset, $\tau_0$, is set to the measured torque, $\tau_m(t)$.
When the active finger moves and the change in torque is larger than the threshold $\tau_t$, a contact is detected (see \cref{fig:pos_torque_contact}).

\section{Experiments}

\begin{figure}[t]
    \centering
	\includegraphics[width=1\linewidth]{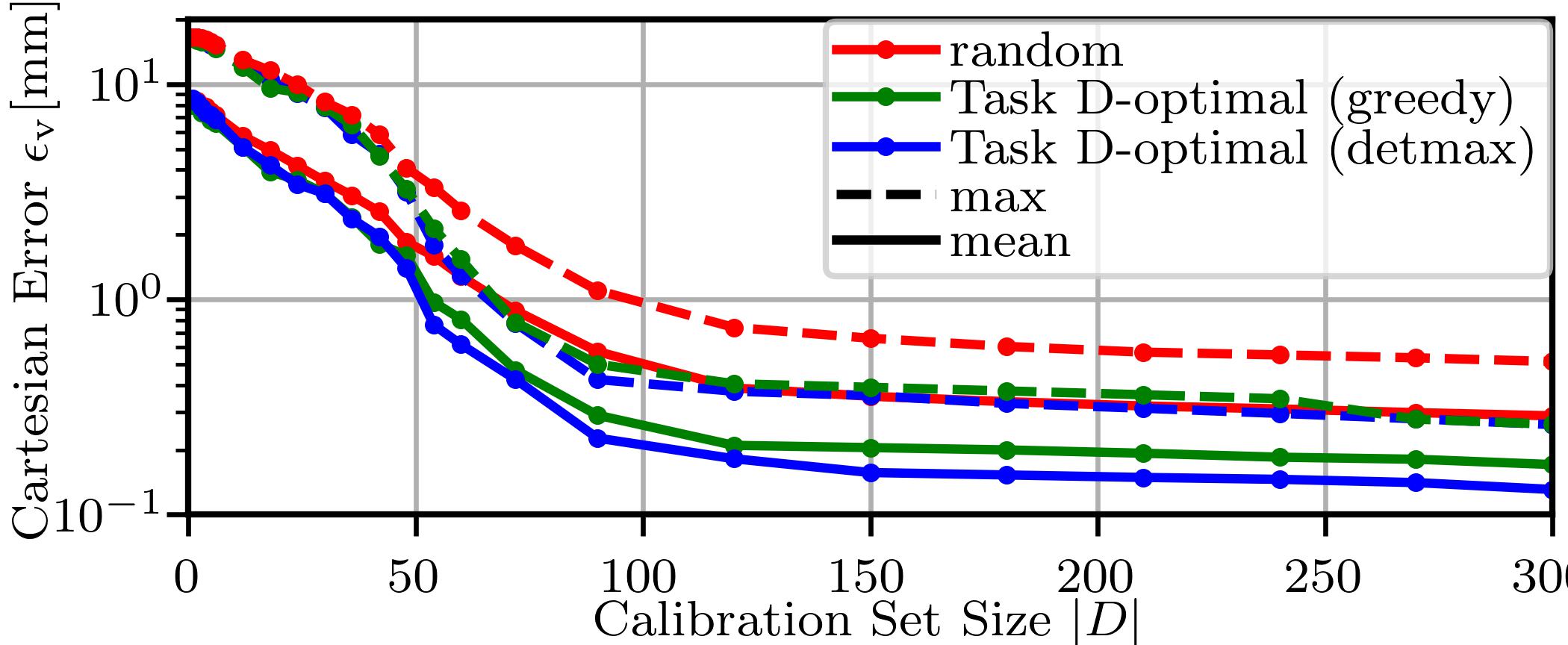}
	\caption{Convergence of mean and maximal cartesian error $\errorVicon$ on the uniform cartesian test set for different calibration sets for the contact measurement function $\funMeasContact$. 
    Randomly chosen sets (red) converge slower than the sets designed according to task D-optimality \cref{eq:optimalityD}.
    The greedy strategy is green, and the detmax\cite{Mitchell2000DOptoimality} strategy is blue.
    The distribution mismatch between calibration and test set can explain why the gap between random and selected samples remains significant even for larger calibration sets.}
    \label{fig:oed_contact}
\end{figure}

As the model of the forward kinematics for the \mbox{DLR-Hand II}, we include all four DH parameters per joint, resulting in $\NTheta = \NDoF \times 4$ calibration parameters.
The hand has four fingers.
Each finger has three active and one passive joint.
This results in $16(=4\times4)$ DH parameters per finger and $64(=4 \times 16)$ parameters for the whole hand.
Furthermore, each finger has three parallel joints, and the ring and middle finger are mounted with the same orientation.

\subsection{Sensitivity Analysis}
To answer the question of whether our contact measurement function $\funMeasContact$ can identify all 
the parameters relevant for the task measurement function $\funMeasTask$, we compute the nullspace of $\JacMeasAll^T\JacMeasAll$ as described in \cref{sec:SensitivityAnalysis}.
We evaluate the respective Jacobians $\JacMeasAllContact$ and $\JacMeasAllTask$ for the nominal robot kinematic.
For the actual measurement function $\funMeasContact$, we use 100 configurations per pair, which were generated as described in \cref{sec:SampleGeneration}.
For the task measurement function $\funMeasTask$, we use 100 configurations per finger drawn from the cartesian test set.

\cref{fig:sensitivity_nominal} shows the results of this analysis for the \mbox{DLR-Hand II}. 
The task measurement function $\funMeasTask$ is drawn in blue, and our contact measurement function $\funMeasContact$ is in red. 
All the pairs are calibrated simultaneously for the rightmost mode ($\bhexagon$). 
Both measurement functions have 56 eigenvalues larger than $1\cdot 10^{-6}$. 
Analyzing the eigenvectors confirms that the kernel of our contact measurement function $\funMeasContact$ is wholly included in the task measurement function $\funMeasTask$ kernel.
Therefore, we can identify all parameters relevant to the task.
The parallel axis of the fingers can explain the $8(=2 \times 4)$ unidentifiable parameters.
Each finger has 3 parallel axes along which a shift can be adjusted without influencing the end-effector, resulting in an increased size of nullspace by 2 per finger and 14 identifiable parameters per finger. 

The leftmost mode ($\bcircle$) shows the calibration of just a single pair. 
Considering the calibration of one pair, from the $32 (=2 \times 16)$ total parameters, $28(=2 \times 14$) should be identifiable. 
However, the critical insight is that measuring between two chains yields less information than measuring between three or more chains.
This holds for the actual scalar distance measurement (red / $\funMeasContact$), where 26 parameters are identifiable, and the task measurement of the relative positions (blue / $\funMeasTask$), where one can identify 27 parameters.  
However, one can measure even less with the scalar distance function when restricting the measurements to a single pair. 
An intuition is that two end-effectors can move on a sphere around each other without changing the scalar distance between them.
Therefore, in this case, one can not identify all the parameters relevant to the task by pure contact measurements.

This invariance generally resolves when adding a third chain ($\btriangle$) to the picture, favoring a holistic calibration of the full kinematic tree. 
\cref{fig:sensitivity_nominal} shows the same analysis but for a generic four-fingered hand without parallel axes or mounting frames.
In theory, all parameters can be identified for the entire hand. 
However, in praxis, eigenvalues below $10^{-6}$ still indicate poor sensitivity. 

\subsection{Optimal Experimental Design}\label{sec:SimulationStudy}
\begin{figure}[t]
    \centering
	\includegraphics[width=1\linewidth]{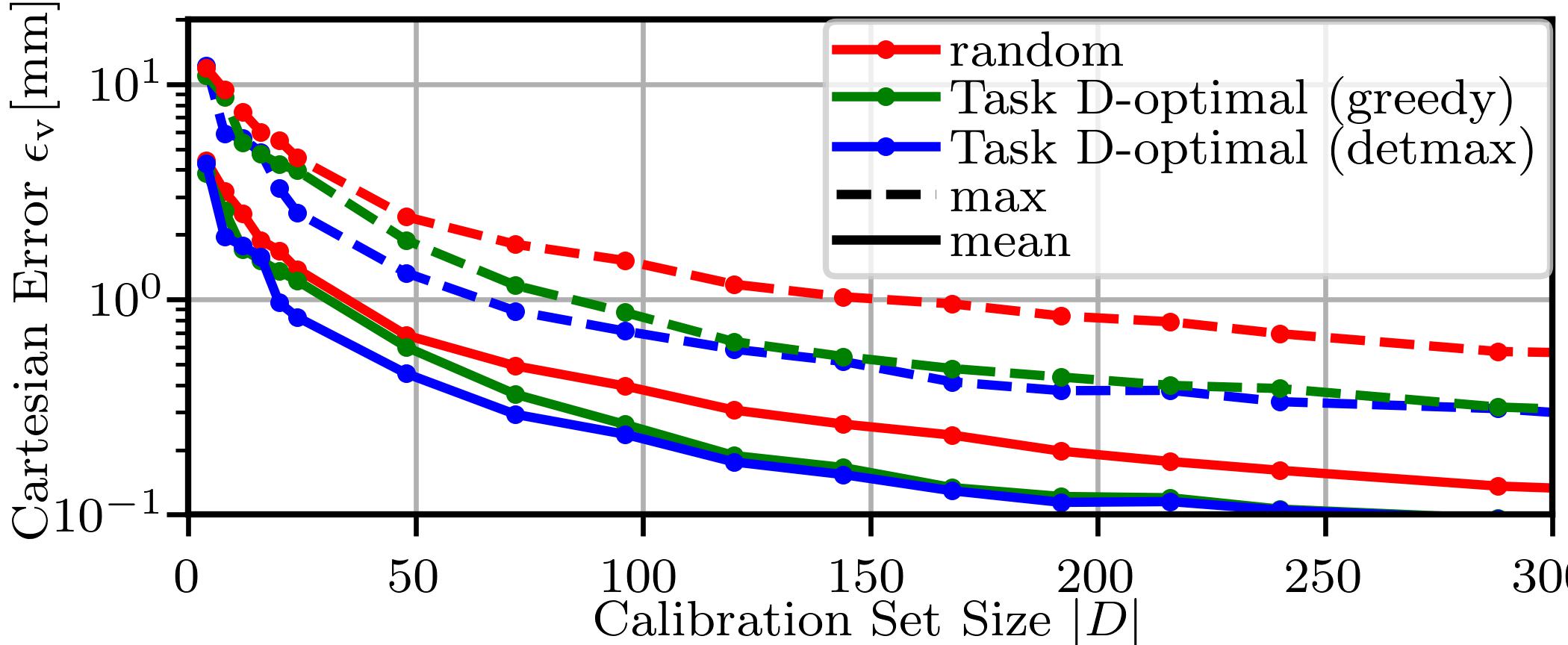}
	\caption{The plot is designed equivalently to \cref{fig:oed_contact}, but for the cartesian measurement function $\funMeasVicon$. 
    The overall convergence of the error $\errorVicon$ is quicker because the absolute cartesian position yields more information per sample.
    Furthermore, the difference between the random approach and the dedicated selection criteria is smaller in this setting.
    One reason is that the calibration set for the external tracking system is closer to the task test set.}
    \label{fig:oed_vicon}
\end{figure}

We conducted an extensive simulation study to verify our analysis in \cref{sec:ProblemAnalysis}.
We use the DH formalism to parametrize the forward kinematics as described in \cref{sec:RobotModel} and apply noise on the nominal DH parameters to obtain new robot models. 
For the rotational DH parameters $\DHalpha$ and $\DHtheta$ we applied sampled uniformly $\pm$ \SI{5}{\degree} and for the translational parameters DH $\DHd$ and $\DHr$ we used uniform noise $\pm$ \SI{5}{mm}.

In this fashion, we created 100 different kinematics to ensure a broad distribution of models. 
Next, we simulated the data collection step and collected measurements for the actual contact measurement function $\funMeasContact$, the actual cartesian measurement function $\funMeasVicon$, and the task measurement function $\funMeasTask$.
On average, the models deviated 21mm from the nominal kinematic on the uniform cartesian test set. 

\cref{fig:oed_contact} show the results of the optimal sample selection introduced in \cref{sec:OED}.
The error on the cartesian test set is drawn over the calibration set size for different selection strategies.
Randomly chosen sets (red) converge slower than the sets designed according to task D-optimality \cref{eq:optimalityD}. 
We compare a greedy strategy (green), which at each step adds the sample $s_i$, which improves the task D-optimality most against the DETMAX algorithm \cite{Mitchell2000DOptoimality} (blue).
This procedure tries to swap samples in an existing calibration set to improve the task D-optimality.
Both selection strategies outperform the random approach, significantly reducing the mean error to 0.1mm with a set of 300 measurements.

The distribution of configurations for the contact measurements differs from the uniform distribution in the cartesian workspace on which we evaluate the calibrations. 
This distribution mismatch can explain why the gap between random and selected samples remains significant even for larger calibration sets.
Our approach using the task D-optimality as a selection criterion does account for this shift directly and improves the contact calibration with its strict constraint for data generation significantly. 
    
\cref{fig:oed_vicon} shows the same analysis but for the cartesian measurement function $\funMeasVicon$ with an external tracking system.
The overall convergence is quicker because the absolute cartesian position yields more information per sample.
Furthermore, the difference between the random approach and the dedicated selection criteria is smaller in this setting.
One reason is that the calibration set for the external tracking system is closer to the task test set.
While proving overall that the selection over task D-optimality is suited to select good calibration sets, those results show that this approach is particularly viable when there is a substantial mismatch between the task measurement function and its test set versus the actual measurement function and its corresponding calibration set.

\begin{table}[t]
    \caption{Calibration Results in mm}
    \label{tab:calibration_results}
    \centering
    \begin{tabular}{c|ccc|c}
\toprule                                                                    & \multicolumn{3}{c|}{Actual Meas. Fun.} & Task Meas. Fun. \\
                                                                    & mean       & std        & max         & mean           \\ \hline
nominal                                                             & 6.07       & 3.90       & 17.70       & 8.01           \\ \cline{1-1}
\begin{tabular}[c]{@{}c@{}}calibrated \\ joint offsets\end{tabular} & 1.04       & 0.82       & 5.13        & 1.36           \\ \cline{1-1}
\begin{tabular}[c]{@{}c@{}}calibrated\\ full DH\end{tabular}        & 0.72       & 0.58       & 3.69        & 0.89  \\         
\bottomrule
\end{tabular}

\end{table}
\begin{figure}[t]
    \centering
	\includegraphics[width=1\linewidth]{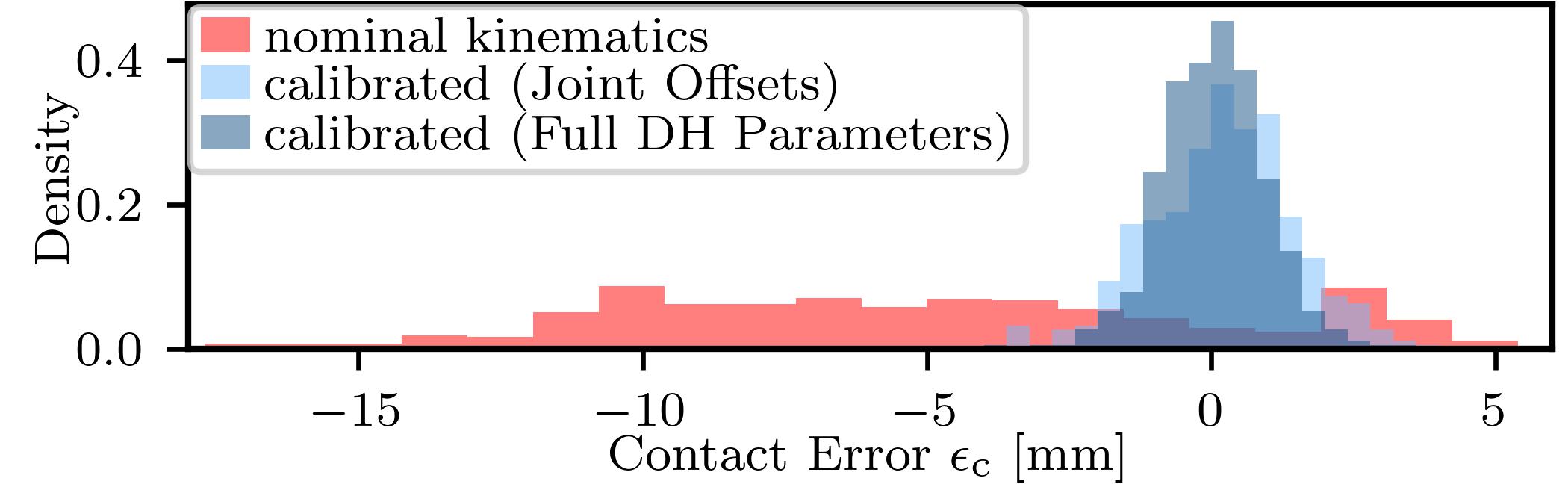}
	\caption{Comparison of contact error $\errorContact$ distribution for the different calibration models. 
             The calibration with all DH parameters can reduce the maximal error to \SI{3.7}{mm}(see also \cref{tab:calibration_results})}.
    \label{fig:error_hist}
\end{figure}

\subsection{Calibration of the real DLR-Hand II}

\begin{figure}[t]
    \centering
	\includegraphics[width=1\linewidth]{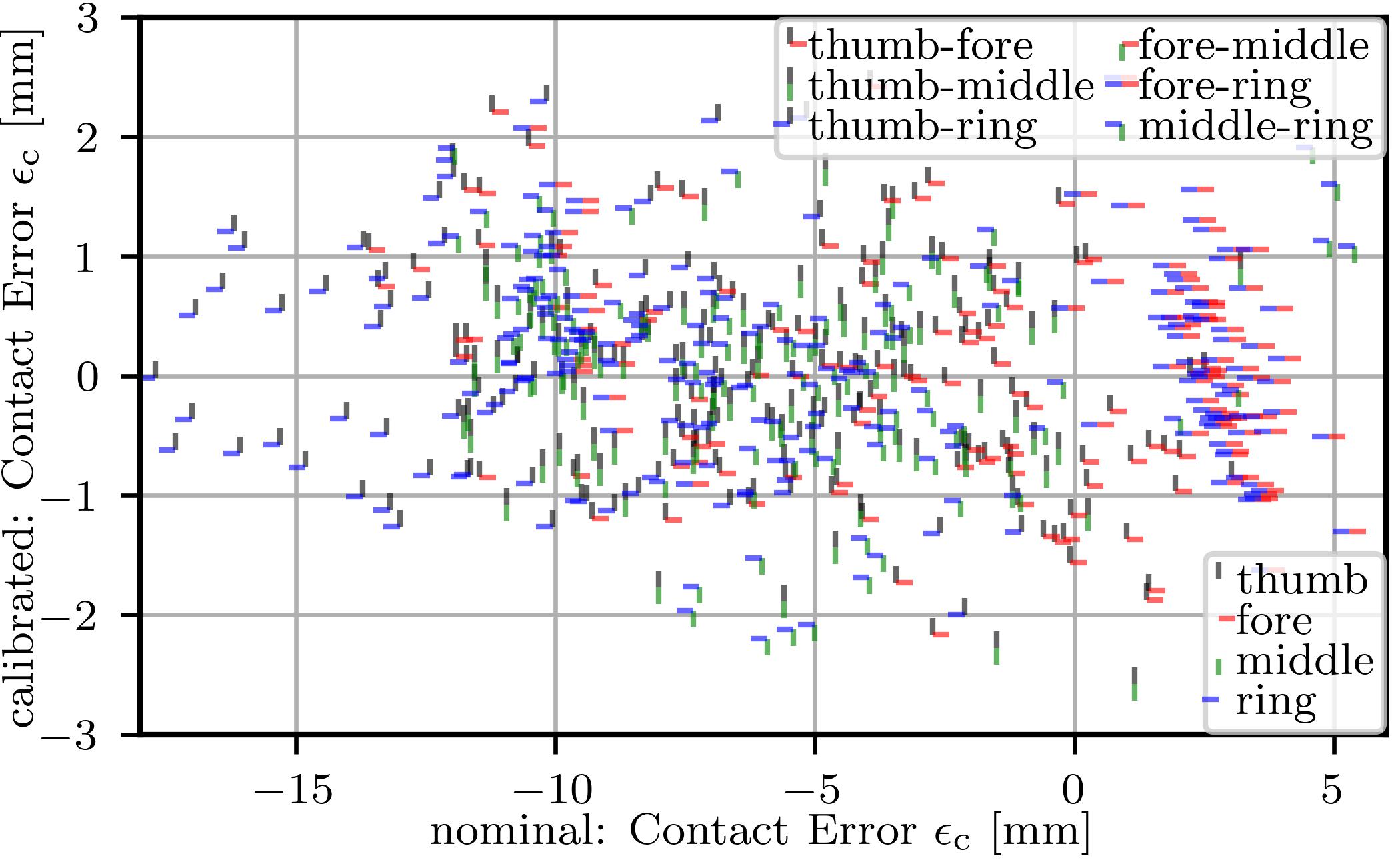}
	\caption{This plot analyses the contact errors $\errorContact$ of the individual fingers and pairs before and after the calibration. 
    Each finger has an individual marker defined by a color and an orientation. 
    A measurement of a specific pair is then an overlay of those two finger markers.
    The x-axis shows the signed distance error before and the y-axis after the calibration, giving detailed insight into the error distribution of the hand.}
    \label{fig:error_finger_before_after}
\end{figure}

Following our methodology described in \cref{sec:ProblemAnalysis} and \cref{sec:SampleGeneration}, we calibrated the DLR-Hand II via pairwise contact measurements.
Overall, we collected 300 samples to ensure we have a large enough set for calibration and evaluation. 
The training-test split was 80/20, and we used cross-validation to get the distribution over the whole dataset. 
Our findings revealed that 150 samples are sufficient for an accurate calibration. 
Collecting that data takes 9 minutes, making this automatic calibration procedure easy and quick to use.

The uncalibrated nominal forwards kinematic has a mean error of \SI{6}{mm} over all 300 samples and a maximal error of up to \SI{17.7}{mm}  for the scalar distance measurement. 
The full error distribution over the signed distance function is shown in red in \cref{fig:error_hist}.
Light blue is the calibrated model using only the joint offsets as calibration parameters, reducing the maximal error to \SI{5.1}{mm}.
The entire calibration model, including all DH parameters, is dark blue. 
This model reduces the maximal error further to \SI{3.7}{mm}.

\cref{fig:error_finger_before_after} analyses the errors of the individual fingers before and after the calibration. 
Each finger has an individual marker, and a measurement of a specific pair overlaps those two finger markers. 
The x-axis shows the signed distance error before and the y-axis after the calibration. 
Besides the significant overall error reduction, one can see that different finger pairs have other error distributions before and after the calibration.
For example, the pair fore-ring is often slightly apart for the nominal model, while the thumb-ring pair is often in deep penetration.

\cref{tab:calibration_results} shows the detailed results of the different models. 
Additionally, the errors in the task space are given.
After calibration, one can use \cref{eq:covariance} to transform the resulting calibration errors of the contact measurement function $\funMeasContact$ into the errors of the task measurement function $\funMeasTask$. 
Starting from the calibration errors $\stndDevMeas$ one can first apply $\JacMeasAllContact$ to map to the uncertainties of the calibration parameters $\cov{\CALpar}$ and then apply in a second step $\JacMeasAllTask$ to map to the task space.   
This results in a reduction of the mean error in the task space form \SI{8}{mm} for the uncalibrated model to \SI{0.9}{mm} after calibration.

\begin{figure}[t]
    \centering
	\includegraphics[width=1\linewidth]{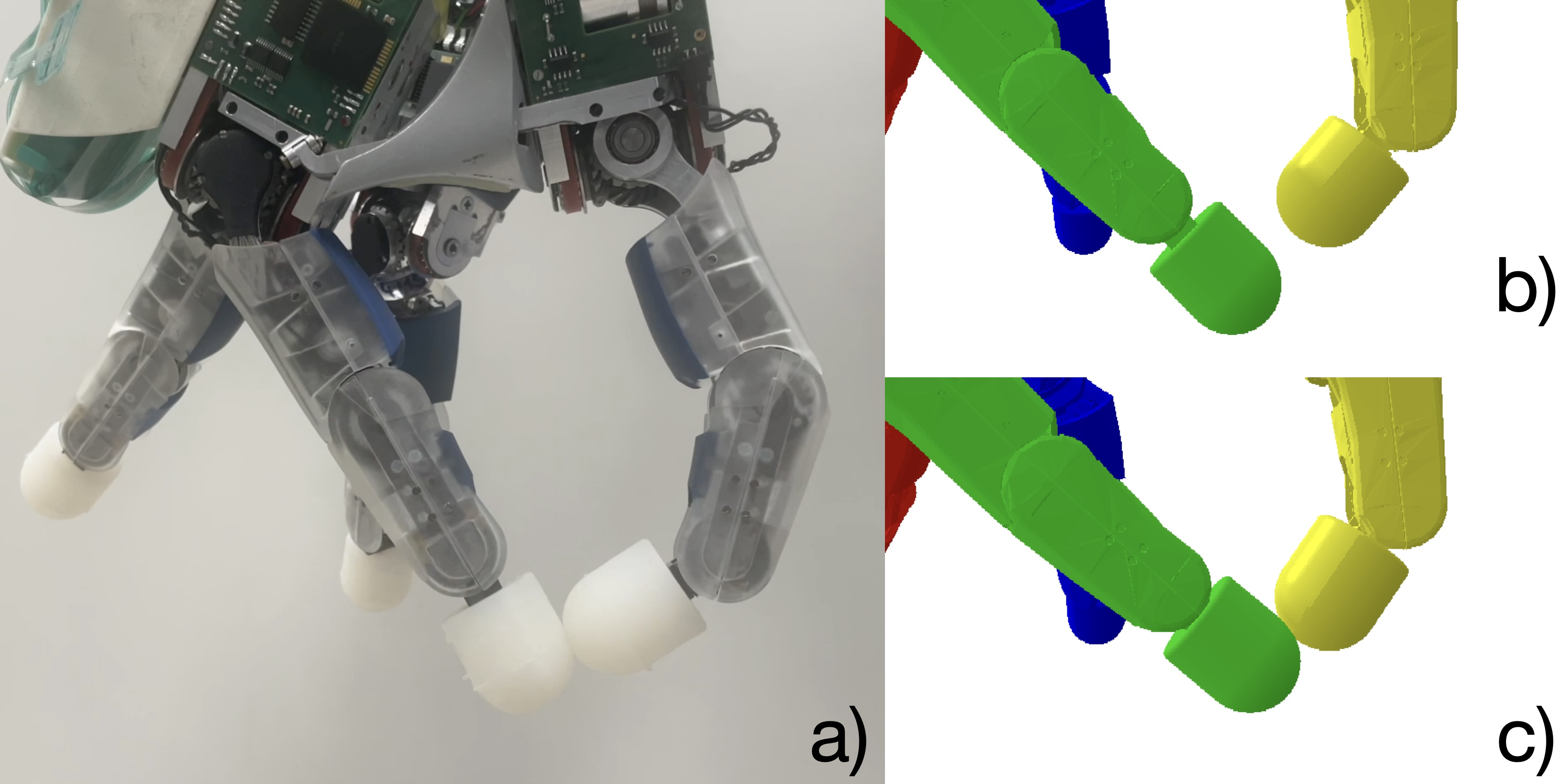}
	\caption{The DLR-Hand II in a finger contact pose from the testset \textit{a)} with the measured joint angles $q$ mirrored to a nominal \textit{b)} and a calibrated \textit{c)} model. 
    The model's error is visibly reduced from a distance of \SI{8.3}{mm} to a penetration of \SI{1}{mm}, allowing dextrous in-hand manipulation.
    }
    \label{fig:calib_vs_noncalib_real_world}
\end{figure}

Finally, \cref{fig:calib_vs_noncalib_real_world} shows the improvement through our calibration procedure on the real DLR-Hand II.
When mirroring a joint configuration $\Joints$, which is in contact, onto the nominal and the calibrated robot model, one can see the sizeable uncalibrated error and the good fit after the calibration, enabling dextrous grasping and in-hand manipulation.

\section{Conclusions and Future Work}\label{sec:Conclusions}

Using a pairwise contact measurement approach, we calibrated the complex robotic DLR-Hand II  with 12 active and 4 passive joints.
This calibration approach has minimal requirements on the robotic hardware and needs no additional tools, making it easy to apply in different setups. 
One only needs a method to contact contacts (e.g., torque sensors in our case) and a mathematical model to describe the distance between the body pairs.

From an uncalibrated distance error of up to \SI{17.7}{mm} (roughly equivalent to $10\%$ of the overall size of the workspace), we could reduce maximal error to \SI{3.7}{mm} and the average error to \SI{0.7}{mm}. 
Looking at the desired task measurement function relevant to dextrous in-hand manipulation, we made a sensitivity analysis to confirm that the scalar distance information between individual finger pairs is enough to identify all relevant DH parameters. 
Furthermore, we showed that performing a joint calibration of multiple finger pairs yields more information than looking at just a single pair.
We used task D-optimality to counteract the mismatch between our desired task measurement function and its corresponding cartesian test set versus the less informative, more constrained contact measurement approach. 
An exhaustive simulation study verifies this selection approach's effectiveness in balancing the calibration and the desired task.

In future work, we want to extend the calibration approach to the robot structure's elasticities, especially in the drivetrain and the fingertips. 
For this, forces are applied via the torque-controlled joints while in contact.

\scriptsize
\bibliographystyle{IEEEtranN-modified}
\bibliography{IEEEabrv, references.bib}
\end{document}